\documentclass{article}

 \usepackage[dblblindworkshop, final]{neurips_2025}

\usepackage[utf8]{inputenc} 
\usepackage[T1]{fontenc}    
\usepackage{hyperref}       
\usepackage{url}            
\usepackage{booktabs}       
\usepackage{amsfonts}       
\usepackage{nicefrac}       
\usepackage{microtype}      
\usepackage{xcolor}         
\usepackage{graphicx}
\usepackage{tabularx,booktabs,array}
\newcolumntype{C}[1]{>{\centering\arraybackslash}p{#1}}
\newcolumntype{L}[1]{>{\raggedright\arraybackslash}p{#1}}
\usepackage[table,xcdraw]{xcolor}
\usepackage{tikz}
\usetikzlibrary{positioning,fit,arrows.meta}

\title{Beyond Generative AI: World Models for Clinical Prediction, Counterfactuals, and Planning}
\workshoptitle{NeurIPS 2025 Workshop on Bridging Language, Agent, and World Models for Reasoning and Planning}

\author{
  Mohammad Areeb Qazi \\
  Mohamed bin Zayed University of Artificial Intelligence (MBZUAI)\\
  Abu Dhabi, UAE \\
  \texttt{mohammad.qazi@mbzuai.ac.ae} \\
  \AND
  Maryam Nadeem \\
  Mohamed bin Zayed University of Artificial Intelligence (MBZUAI)\\
  Abu Dhabi, UAE \\
  \texttt{maryam.nadeem@mbzuai.ac.ae} \\
  \AND
  Mohammad Yaqub \\
  Mohamed bin Zayed University of Artificial Intelligence (MBZUAI)\\
  Abu Dhabi, UAE \\
  \texttt{mohammad.yaqub@mbzuai.ac.ae} \\
}

\begin{document}

\maketitle

\begin{abstract}
    Healthcare requires AI that is predictive, reliable, and data-efficient. However, recent generative models lack physical foundation and temporal reasoning required for clinical decision support. As scaling language models show diminishing returns for grounded clinical reasoning, \emph{world models} are gaining traction because they learn multimodal, temporally coherent, and action-conditioned representations that reflect the physical and causal structure of care. This paper reviews \emph{World Models} for healthcare systems that learn predictive dynamics to enable multistep rollouts, counterfactual evaluation and planning. We survey recent work across three domains: (i) medical imaging and diagnostics (e.g., longitudinal tumor simulation, projection-transition modeling, and Joint Embedding Predictive Architecture i.e., JEPA-style predictive representation learning), (ii) disease progression modeling from electronic health records (generative event forecasting at scale), and (iii) robotic surgery and surgical planning (action-conditioned guidance and control). We also introduce a capability rubric: \textbf{L1} temporal prediction, \textbf{L2} action-conditioned prediction, \textbf{L3} counterfactual rollouts for decision support, and \textbf{L4} planning/control. Most reviewed systems achieve \textbf{L1--L2}, with fewer instances of \textbf{L3} and rare \textbf{L4}. We identify cross-cutting gaps that limit clinical reliability; under-specified action spaces and safety constraints, weak interventional validation, incomplete multimodal state construction, and limited trajectory-level uncertainty calibration. This review outlines a research agenda for clinically robust \emph{prediction-first} world models that integrate generative backbones (transformers, diffusion, VAE) with causal/mechanical foundation for safe decision support in healthcare.
\end{abstract}

\section{Introduction}

Healthcare systems are under tremendous pressure from aging populations, chronic diseases, and shortages of workers. Populations are rapidly aging (projected 1 in 6 over 60 by 2030)~\cite{world2015world}, and global health workers are estimated to exceed 10 million by 2030~\cite{world2015world}. This results in ever-growing clinical data, underscoring the importance of AI-driven solutions to improve the quality and efficiency of care. During the past decade, AI in healthcare has evolved from traditional statistical models to deep learning and, more recently, large generative models. Deep neural networks delivered breakthroughs in medical imaging and diagnostics, leading to a broader use in fields such as radiology, pathology, and similar areas\cite{litjens2017survey,esteva2017dermatology,gulshan2016dr,areeb2022helping,areeb2021deep,qazi2023multi}. Large Language Models (LLMs) based on transformer architectures now demonstrate strong performance on clinical NLP tasks (e.g., near-expert USMLE performance)~\cite{rabbani2025generative}. In parallel, image-generative models have emerged as powerful tools; diffusion models synthesize realistic MRI/CT for augmentation and anonymization~\cite{khader2023denoising,hashmi2024xreal,alam2024introducing}, while VAEs and related frameworks generate synthetic patient data across modalities~\cite{ibrahim2025generative}. These advances illustrate how transformers, diffusion models, and VAEs are reshaping research and practice, enabling improved prediction, data augmentation, and discovery.

\begin{figure}%
    \centering
\includegraphics[width=\textwidth]{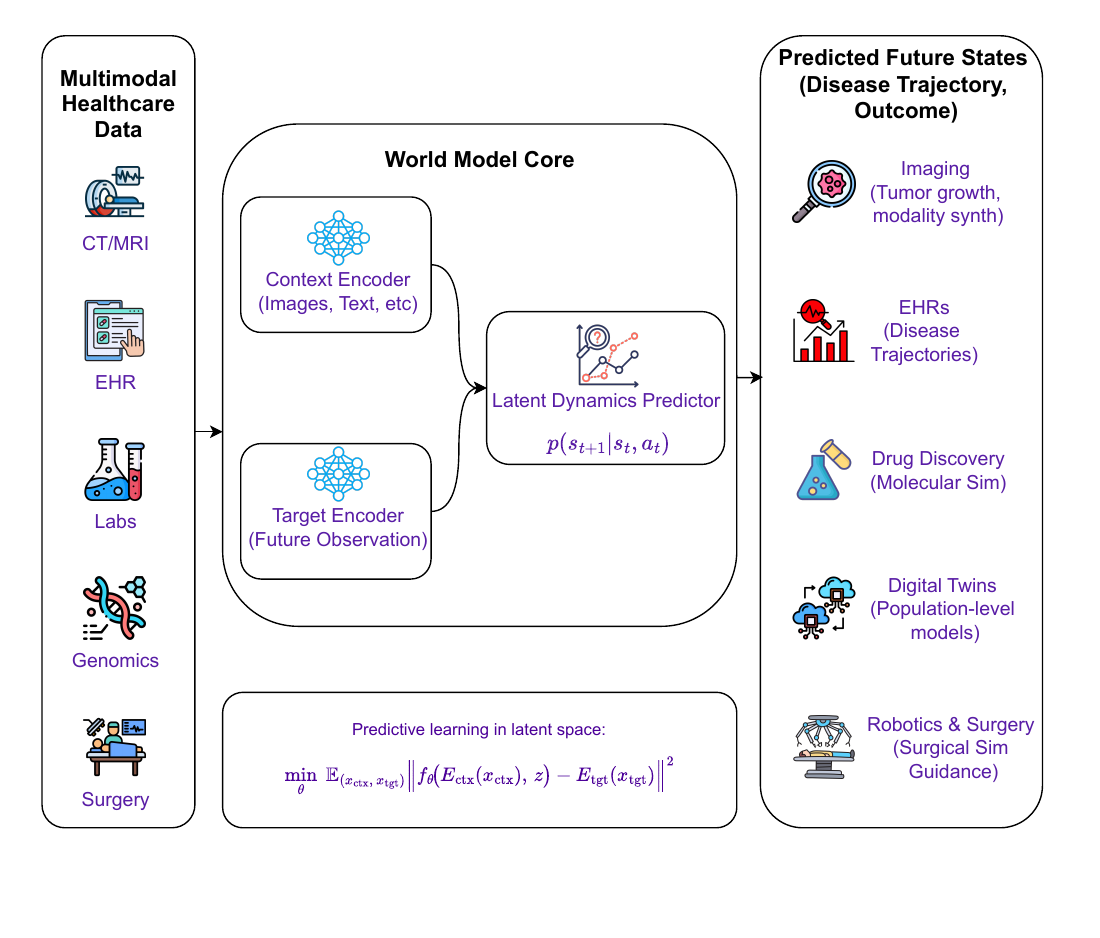} 
    \caption{Conceptual schematic of world models for healthcare. Multimodal clinical inputs are encoded into a latent state; a latent dynamics predictor models transitions \(p(s_{t+1}\mid s_t, a_t)\); predicted futures support downstream tasks across imaging, EHR trajectories, drug discovery, surgical robotics, and digital twins. The JEPA-style objective (bottom) trains a predictor \(f_{\theta}\) to map context embeddings \(E_{\mathrm{ctx}}(x_{\mathrm{ctx}})\) and latent dynamics \(z\) to match target embeddings \(E_{\mathrm{tgt}}(x_{\mathrm{tgt}})\).}
    \label{fig:wm_main}
\end{figure}

However, generative models are insufficient for high-stakes medicine. They lack grounding in the physical, spatial, and causal structure of clinical reality, and can produce plausible but incorrect outputs (“hallucinations”) with dangerous consequences~\cite{rabbani2025generative}. This motivates a prediction-first, world-based alternative in which the model learns \emph{predictive dynamics} - often formalized as \(p(s_{t+1}\mid s_t,a_t)\) or via future-latent predictive objectives (e.g. JEPA)~\cite{lecun2022path}. In this paradigm, a \emph{world model} (WM) is an explicit generative model of state dynamics that supports internal simulation, counterfactual evaluation, and planning. Early works show that agents can learn compact latent 'worlds' and train policies inside them before successful transfer to the real world~\cite{ha2018world}, and recent medical AI has begun to explore such ideas in clinical imaging and surgical simulation~\cite{koju2025surgical}.

This paper presents the first focused review of \emph{world models in healthcare}, covering medical imaging and diagnostics, disease progression modeling, and robotic surgery/surgical planning. To compare heterogeneous methods, we introduce a capability rubric: \textbf{L1} temporal prediction; \textbf{L2} action-conditioned prediction; \textbf{L3} counterfactual rollouts for decision support; \textbf{L4} planning/control. We analyze how these works adapt concepts from generative modeling and model-based reasoning, assess the extent of current progress, and identify open challenges. Because WMs in medicine are still in an early phase, a systematic synthesis can guide research efforts, highlight opportunities, and clarify the trajectory of this emerging area of clinical AI.


\section{Background and Foundations}

A WM in machine learning refers to a system that learns an internal generative model of the dynamics of the environment, enabling the prediction of future states given current observations and potential actions. Formally, a WM approximates the transition distribution $p(s_{t+1} \mid s_t, a_t)$, where $s_t$ denotes the state at time $t$ and $a_t$ is an action taken by the agent as shown in figure \ref{fig:wm_main}. This contrasts with discriminative or purely generative models that focus on classification or static sample generation; WMs emphasize learning predictive dynamics that allow simulation, counterfactual reasoning, and planning. The idea has roots in reinforcement learning (RL), most notably in Sutton’s Dyna architecture \cite{sutton1990integrated}, which combined real experience with updates from a learned model to accelerate policy learning. The concept was revitalized in deep learning through a WM framework \cite{ha2018recurrent}, demonstrating that compact latent representations of environments, learned with generative recurrent neural networks, could serve as internal simulators where agents learn policies 'in their dreams'. 

Since 2018, WM research has expanded significantly, with architectures such as Dreamer \cite{hafner2019learning}, SimCore \cite{gregor2019shaping}, and MuZero \cite{schrittwieser2020mastering} demonstrating the advantages of model-based reasoning in complex and high-dimensional domains. More recently, the Joint Embedding Predictive Architecture (JEPA) has been proposed as a unifying framework for predictive learning \cite{lecun2022path}. JEPA emphasizes learning representations by predicting future latent embeddings rather than raw observations, making it particularly suitable for high-dimensional data such as video or multimodal medical records. These developments reflect a broader trajectory: from early model-based RL with hand-crafted features, to latent-variable generative models capable of internal simulation and planning.

WMs are closely linked with advances in large generative architectures. Variational autoencoders (VAE) \cite{kingma2013auto} enable compression of high-dimensional inputs (e.g., imaging data) into structured latent spaces that can be rolled forward in time. Diffusion models \cite{ho2020denoising} offer powerful generative capabilities with uncertainty quantification, increasingly applied to imaging tasks, including the synthesis of medical data. Transformers \cite{vaswani2017attention} excel at modeling long-range dependencies and multimodal fusion, making them natural backbones for WMs in clinical settings where data span temporal, textual, and visual modalities. In this sense, WMs can be viewed as an extension of generative AI. Instead of generating only text or images, they generate trajectories of states and outcomes, providing a predictive simulator for decision support.

Healthcare is a particularly compelling domain for WMs. Clinical problems are inherently multimodal, temporal, and causal: disease trajectories unfold over time, interventions alter patient states, and data sources range from imaging to electronic health records (EHRs) to genomics. Traditional discriminative models, even large generative ones, struggle to capture these dynamics in a way that supports clinical reasoning. WMs can integrate multimodal patient data into a unified latent state, simulate how that state evolves under different interventions, and thus enable critical what-if analyses for decision making. By grounding predictions in learned representations of medical reality, WMs have the potential to overcome the brittleness and lack of physical grounding observed in current generative AI systems \cite{lecun2022path,lecun2024beyond}.

\section{Findings and Discussion}

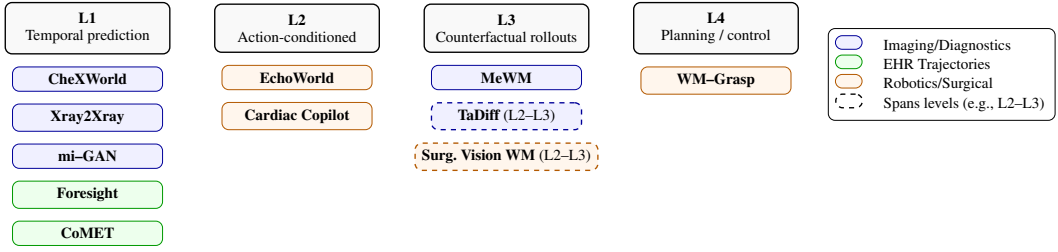
\begin{figure}[t]
\centering
\resizebox{\columnwidth}{!}{%
\begin{tikzpicture}[
  font=\footnotesize,
  >=Latex,
  col/.style={draw, rounded corners, thick, align=center, inner sep=6pt, minimum width=3.4cm, fill=gray!5},
  paper/.style={draw, rounded corners=4pt, align=center, inner sep=4pt, minimum width=3.1cm},
  imaging/.style={paper, draw=blue!60!black, fill=blue!6},
  ehr/.style={paper, draw=green!60!black, fill=green!7},
  robot/.style={paper, draw=orange!70!black, fill=orange!7},
  span/.style={dashed, line width=0.8pt}
]

\node[col] (L1) at (0,0) {\textbf{L1}\\Temporal prediction};
\node[col, right=0.9cm of L1] (L2) {\textbf{L2}\\Action-conditioned};
\node[col, right=0.9cm of L2] (L3) {\textbf{L3}\\Counterfactual rollouts};
\node[col, right=0.9cm of L3] (L4) {\textbf{L4}\\Planning / control};

\def\dy{0.25}

\node[imaging, below=\dy of L1.south] (chex) {\textbf{CheXWorld}};
\node[imaging, below=\dy of chex.south] (x2x) {\textbf{Xray2Xray}};
\node[imaging, below=\dy of x2x.south] (migan) {\textbf{mi--GAN}};
\node[ehr, below=\dy of migan.south] (fore) {\textbf{Foresight}};
\node[ehr, below=\dy of fore.south] (comet) {\textbf{CoMET}};

\node[robot, below=\dy of L2.south] (echo) {\textbf{EchoWorld}};
\node[robot, below=\dy of echo.south] (copilot) {\textbf{Cardiac Copilot}};

\node[imaging, below=\dy of L3.south] (mewm) {\textbf{MeWM}};
\node[imaging, span, below=\dy of mewm.south] (tadiff) {\textbf{TaDiff} (L2--L3)};
\node[robot, span, below=\dy of tadiff.south] (surgwm) {\textbf{Surg.\ Vision WM} (L2--L3)};

\node[robot, below=\dy of L4.south] (wmgrasp) {\textbf{WM--Grasp}};

\node[draw, rounded corners, inner sep=4pt, align=left, right=0.6cm of L4.east, anchor=north west] (legend) {
\begin{tabular}{@{}ll@{}}
\tikz{\node[imaging, minimum width=0.55cm, minimum height=0.28cm]{};} & Imaging/Diagnostics \\
\tikz{\node[ehr, minimum width=0.55cm, minimum height=0.28cm]{};}     & EHR Trajectories \\
\tikz{\node[robot, minimum width=0.55cm, minimum height=0.28cm]{};}   & Robotics/Surgical \\
\tikz{\node[paper, span, minimum width=0.55cm, minimum height=0.28cm]{};} & Spans levels (e.g., L2--L3) \\
\end{tabular}
};

\end{tikzpicture}%
}
\caption{Capability map of reviewed papers across four levels: L1 (temporal prediction), L2 (action-conditioned prediction), L3 (counterfactual rollouts for decision support), L4 (planning/control). Colors denote domains (Imaging, EHR, Robotics). Dashed borders indicate methods spanning adjacent levels (e.g., TaDiff and Surgical Vision WM at L2--L3).}
\label{fig:capability-map}
\end{figure}

Recently, several studies have pushed toward world-model–style learning in healthcare. For clarity, we group the literature by application area: Medical Imaging and Diagnostics, Disease Progression Modeling (EHR), and Robotic Surgery \& Surgical Planning, and briefly summarize each study's contribution in Table \ref{tab:wm_papers}.

\textbf{Medical Imaging and Diagnostics}: Imaging is a natural testbed for learned dynamics because anatomical states evolve and interventions alter future observations. Medical World Model (MedWM) formulates treatment planning as an action-conditioned simulation: A vision language policy proposes interventions and a generative dynamics model predicts posttreatment tumor states for protocol selection \cite{yang2025mewm}. Mi-GAN predicts future 3D brain MRI volumes to model Alzheimer’s progression from a baseline scan using a multi-information GAN \cite{zhao2021migan}. CheXWorld uses a JEPA-style predictive objective with context/target encoders and a predictor to learn radiograph representations that capture local anatomy, global layout, and domain variation \cite{yue2025chexworld}. Xray2Xray learns projection-transition dynamics at all acquisition angles to encode latent 3D chest volume for downstream risk and diagnostic tasks \cite{yang2025xray2xray}. Treatment-aware diffusion (TaDiff) conditions longitudinal MRI generation and tumor segmentation in therapy to forecast diffuse glioma evolution under alternative treatment plans \cite{liu2023tadiff}.

\textbf{Disease Progression Modeling}: Longitudinal EHRs are event streams in which generative models simulate trajectories and forecast future outcomes. Foresight is a generative transformer that converts clinical text to coded concepts and auto-regressively forecasts future disorders, procedures, substances, and findings in large hospital cohorts \cite{kraljevic2024foresight}. Generative Medical Event Models Improve with Scale (CoMET) trains decoder-only transformers on billions of medical events (Epic Cosmos), showing scaling laws and improved multitask predictions while simulating patient-timeline event sequences \cite{waxler2025comet}.

\textbf{Robotic Surgery \& Surgical Planning}: Guidance and control require models that couple the visual state with action-conditioned dynamics. EchoWorld pre-trains a motion-aware WM for echocardiography that encodes anatomy and the effect of probe motion, reducing plane-guidance error on large-scale ultrasound data \cite{yue2025echoworld}. Cardiac Copilot introduces a WM 'Cardiac Dreamer' whose latent spatial features provide a navigation map for real-time probe guidance, improving navigation error on clinical scans \cite{jiang2024cardiaccopilot}. Surgical Vision World Model (SurgWM) adapts action-controllable video generation (latent action inference + dynamics) to synthesize controllable surgical video from unlabeled data \cite{koju2025surgicalwm}.


Figure~\ref{fig:capability-map} summarizes the distribution of the reviewed works along a capability ladder from \textbf{L1} (temporal prediction) to \textbf{L4} (planning/control). The literature is concentrated at \textbf{L1--L2}: radiograph and longitudinal MRI papers emphasize future prediction in observation or latent space (e.g., future-latent predictive objectives or generation conditioned by therapy), while EHR models largely auto-regress event sequences without explicit actions. \textbf{L3--L4} capabilities counterfactual rollouts used for decision support and closed-loop control are comparatively rare, emerging primarily in treatment-aware imaging simulators and robotic/surgical systems. This pattern aligns with the arguments that clinically useful systems must move from sample generation to \emph{prediction-first, world-grounded} modeling of dynamics and interventions (for example, learning \(p(s_{t+1}\mid s_t,a_t)\) or future latent prediction) rather than language-only next-token modeling~\cite{lecun2024beyond,lecun2022path}.

Domain effects are visible in the map. \textbf{Imaging/diagnostics} spans L1--L3: predictive representation learning supports transfer (L1), projection/longitudinal dynamics enable interventional ``what-if'' simulation (L2--L3), yet few studies close the loop with planning. \textbf{EHR} work concentrates at L1, reflecting strong timeline forecasting on scale but limited action semantics and limited counterfactual validation. \textbf{Robotics/surgical planning} naturally advances to L2 and beyond because actions (probe/tool motion) are explicit; instances of L3 appear where controllable video generation is used for simulated outcomes and L4 is reached when learned dynamics drive control policies. Across domains, large generative backbones (transformers, diffusion, VAE) are common, but the defining step toward ``world modeling'' is the \emph{use of learned dynamics for rollouts under interventions}---not generative fluency per se.

Two cross-cutting gaps explain why many methods remain at L1--L2. First, \textbf{action semantics and constraints} are underspecified outside robotics: therapy, protocol, dose, or timing are rarely formalized as action variables with units and safety limits. Second, \textbf{ the interventional validity} is weakly measured: simulated futures may be realistic, but incorrect about treatment effects. Progress on these fronts, together with multimodal state construction, trajectory-level uncertainty calibration, and decision-aligned evaluation, will be decisive in advancing toward clinically reliable L3--L4 WMs~\cite{lecun2024beyond}.

\begin{table}[t]
\caption{Representative works at the intersection of large generative models and world modeling in healthcare. Short names are shown in bold; citations refer to the bibliography.}
\centering
\scriptsize
\setlength{\tabcolsep}{2pt}
\renewcommand{\arraystretch}{1.1}
\begin{tabularx}{\columnwidth}{@{}C{0.85cm}L{2.4cm}L{2.4cm}X@{}}
\toprule
\rowcolor[HTML]{F2F2F2} 
\textbf{Year} & \textbf{Paper} & \textbf{Category} & \textbf{Description} \\
\midrule
2025 & \textbf{MeWM} \cite{yang2025mewm} & Imaging \& Planning & Action-conditioned 3D generator simulates post-treatment tumor from CT/EHR for protocol selection. \\
2025 & \textbf{CoMET} \cite{waxler2025comet} & EHR Trajectories & Scalable generative event model forecasting multi-horizon clinical timelines. \\
2021 & \textbf{mi\textendash GAN} \cite{zhao2021migan} & Imaging (Neuro) & Multi-information GAN predicts future 3D MRI to model Alzheimer’s progression. \\
2025 & \textbf{CheXWorld} \cite{yue2025chexworld} & Radiography & JEPA-style latent prediction for local anatomy, global layout, and domain shifts. \\
2025 & \textbf{EchoWorld} \cite{yue2025echoworld} & Ultrasound Guidance & Motion-aware world modeling under probe motion improves echo plane guidance. \\
2025 & \textbf{Surg-VM} \cite{koju2025surgicalwm} & Surgical Video Sim & Action-controllable surgical video generation with latent actions for training. \\
2024 & \textbf{WM-Grasp} \cite{lin2024wmgrasp} & Surgical Grasping & World-model RL for general surgical grasping robust to object variation/disturbance. \\
2024 & \textbf{Cardiac Copilot} \cite{jiang2024cardiaccopilot} & Ultrasound Guidance & World-model (“Cardiac Dreamer”) encodes cardiac spatial structure for probe navigation. \\
2025 & \textbf{Xray2Xray} \cite{yang2025xray2xray} & Radiography (Vol) & World model learns 3D volumetric context by modeling projection transitions. \\
2023 & \textbf{TaDiff} \cite{liu2023tadiff} & Neuro MRI & Treatment-aware diffusion predicts longitudinal MRIs and tumor masks under therapies. \\
2024 & \textbf{Foresight} \cite{kraljevic2024foresight} & EHR Forecasting & Generative transformer forecasting future medical events from patient timelines. \\
\bottomrule
\end{tabularx}
\label{tab:wm_papers}
\end{table}

\section{Future Work}
\label{sec:future-work}

Although world–modeling approaches are beginning to deliver predictive representations and simulation capabilities in healthcare, advances are needed in various aspects. Here, we present the gaps and outline directions to address them.

\begin{itemize}
    \item \textbf{From L1 to L2: formalize actions.}  
    Define clinically meaningful action spaces per domain: imaging (protocol/angle/contrast), EHR (drug, dose, timing), robotics (tool pose / force) ---- and encode safety constraints within the model and interface of implementation.

    \item \textbf{From L2 to L3: establish counterfactual correctness.}  
    Evaluate ``what-if'' rollouts for interventional validity via multi-site holdouts, matched cohorts or natural experiments, and clinician adjudication; favor decision-centric endpoints (e.g., survival, adverse events) over perceptual or token-level metrics.

    \item \textbf{From L3 to L4: close the loop.}  
    Couple learned dynamics with model-based RL/MPC and reliable off-policy evaluation; stage prospective, small-$N$ pilots prior to deployment, with uncertainty-aware planning and explicit abstention policies.

    \item \textbf{Multimodal integrated state at scale.}  
    Learn unified latent states fusing imaging, EHR, waveforms, and genomics under missingness and irregular sampling; characterize scaling laws linking data/model size to rollout fidelity and decision value.

    \item \textbf{Trajectory-level uncertainty and robustness.}  
    Provide calibrated predictive distributions over entire trajectories (e.g., conformal bands for longitudinal imaging/EHR), stress-test out-of-distribution shifts (scanner/site, demographics), and assess subgroup fairness of \emph{recommended actions}.

    \item \textbf{Causal and mechanistic grounding.}  
    Combine learned dynamics with causal identification (adjustment sets, instrumental variables, transportability checks) and embed mechanistic priors (projection geometry, tumor growth or PK--PD, anatomy/physics) to improve extrapolation and safety.

    \item \textbf{Evaluation and reporting standards.}  
    Adopt shared capability benchmarks aligned with the ladder in Figure~\ref{fig:capability-map}; standardize horizon lengths, action definitions, counterfactual protocols, and ablations isolating action-conditioning and planning components.

    \item \textbf{Tooling, privacy, and governance.}  
    Release reference implementations and lightweight simulators (imaging acquisition, EHR counterfactual sandboxes, surgical video) for reproducibility; define governance for simulation-based recommendations with privacy-preserving training and audited synthetic data.

    \item \textbf{Efficiency and deployability.}  
    Use distillation, amortized planning, and latent-space MPC for real-time robotics and bedside decision support; report latency, compute, and energy alongside accuracy.
\end{itemize}

\section{Conclusion}
\label{sec:conclusion}

This review aimed to clarify what constitutes a \emph{world model} for healthcare and review recent work within a capability-oriented rubric. We distinguished world modeling from generic generative AI by its focus on \emph{predictive dynamics}, often formalized as \(p(s_{t+1}\mid s_t,a_t)\) or future-latent prediction 
that enable rollouts, counterfactual evaluation, and planning. Surveying diagnostic imaging, EHR disease progression, and robotic surgery, we found that most methods currently achieve \textbf{L1--L2} capabilities (temporal prediction, action-conditioned prediction), with fewer instances of \textbf{L3} counterfactual decision support and rare \textbf{L4} closed-loop planning/control. The field has focused around large generative backbones (transformers, diffusion, VAEs) paired with latent dynamics learning (e.g., JEPA-style objectives), but the defining ingredient is the \emph{use of learned dynamics} to reason under interventions rather than generative fluency alone.

Our synthesis highlights the central obstacles to progress: underspecified clinical \emph{action spaces} (and associated safety constraints) outside robotics; limited evaluation of \emph{interventional validity} for simulated futures; incomplete construction \emph{multimodal state} in EHR, imaging and signals; and lack of standardized, decision-aligned evaluation protocols. Addressing these gaps is essential for moving beyond accurate forecasting toward reliable counterfactual reasoning and clinically actionable planning. WMs that integrate principled action semantics, calibrated trajectory uncertainty, and causal/mechanistic grounding can elevate healthcare AI from pattern recognition to faithful simulation and safe decision support. As community benchmarks converge on a capability ladder (L1--L4), and as open tooling, governance, and privacy-preserving training mature, we anticipate a shift toward \textbf{L3--L4} systems that meaningfully assist clinicians: simulating treatment alternatives, guiding imaging and procedures, and closing the loop in safety-critical settings. Realizing this vision will require sustained collaboration across machine learning, clinical science, and health systems engineering; the works reviewed here provide the scaffolding on which the next generation of clinically robust WMs will be built.

\bibliographystyle{unsrt}
\bibliography{references}

\end{document}